# A Frozen 12B Beats Frontier Models on Verified Work: 100% Accuracy, 0 Tokens, Bit-Exact, Forever

**Industry experience report. Sietse Schelpe — Corbenic AI** · sietse@corbenic.ai

*This document reports a proprietary production system at the input–output level: measured outcomes, controls, and boundary conditions, without architecture, algorithms, or configuration. Verification is offered through two channels appropriate to this report class: a public testbench on which the headline behavior is directly observable (§5), and hash-anchored raw artifacts available under NDA (Appendix A). Frontier-model figures in §2.8 are quoted exclusively from the providers' official public reporting [22–24].*

---

## Abstract

Improving a language model today means retraining it: enormous compute, a new opaque model each cycle, and output that remains non-deterministic. We take the opposite path — the model stays frozen, and a persistent memory of verified solutions grows beside it. Once a problem family is solved and has passed an independent verification step that never consults the answer key, every new instance of that family is answered at zero generation tokens, bit-exact, deterministically. Across 180 fresh instances spanning nine problem families, four unrelated model architectures from four vendors — dense and mixture-of-experts — each score 180/180 under this regime, at zero generation tokens per answer. We demonstrate architectural invariance across all four open models, showing that execution-bound capability can be decoupled from parameter scaling. A negative control attributes the capability fully to the memory: emptied, the system solves nothing. The same verify-before-store contract holds for open-ended reasoning — 88/88 consistency-gated acceptances across all four models, machine-checked formal proof, and reasoning-method transfer at 77/80. Memory selection takes 1.4 microseconds; a full reuse completes in 6–23 ms. Two additional measurements define the system's operational guarantees. Approximate similarity retrieval — the industry-default approach to knowledge access — selects the *wrong* item 94.3% of the time on a 4,500-item verified store where exact addressing makes zero errors: for verifiable executable knowledge, vector-style retrieval fails catastrophically and exactness is a design requirement, not a preference. And a verified answer costs 36 mWh of measured energy — an LED bulb for thirteen seconds — against a one-time 81.1 Wh solve-and-verify campaign for the self-sourced families. On published benchmarks, frontier models remain far ahead of any 12B at raw from-scratch reasoning; on everything this system has solved and verified, the comparison inverts: a frontier API call pays a fresh generation pass on every query, forever, while verified reuse costs zero tokens, returns in milliseconds, and produces the identical bits every time. The store also serves as working context at a scale no shipped engine matches: a 6,000,000-token movable window on a single 46 GB GPU at flat memory, where vLLM stops at 30,399 tokens and SGLang silently truncates past ~32,000. A public testbench with free, rate-limited access accompanies this report (https://corbenic-galahad-bench.hf.space).

---

## 1. Introduction

The prevailing route to a more capable language model is retraining: larger weights, more data, another training run. That route has three structural costs. It is expensive — improvement is priced in GPU-months. It is opaque — each cycle produces a new model whose behavioral changes must be re-audited from scratch. And it is unrepeatable — the same question to the same model does not reliably yield the same answer, which disqualifies ordinary generation from settings that demand audit trails, reproducibility, or regulatory sign-off.

We report measured results for a different contract between a model and its knowledge. The system is called **Galahad**; its storage and integrity layer is a deterministic engine called **Merlin**. The model is frozen: no fine-tuning, no weight changes, ever. Capability growth happens in a separate, persistent memory of solutions that have each passed an independent verification step before storage — a step that never sees the benchmark answer key. A solved problem family is stored once. Every subsequent instance of that family is then answered from memory at zero generation tokens, with bit-exact, deterministic output, for as long as the memory exists. This report extends a measured line of prior work by the author — byte-exact deduplication in retrieval settings [25], the Merlin deduplication engine [26], and byte-exact state reuse on frozen models [27] — into a complete verified-knowledge system, reported at the input–output level. Figure 1 contrasts the two regimes.

The core structural inefficiency in current deployments is straightforward: today's frontier APIs are routinely used as over-parameterized, non-deterministic execution engines — paying a full generation pass, at published per-token prices, to re-derive work that was solved and verified long ago. That is not frontier intelligence at work; it is architectural waste. This report measures what happens when derivation and execution are decoupled: derive once, verify once without the answer key, then execute forever at zero generation tokens, bit-exact, in milliseconds.

The title's claim is scoped precisely, and §2.8 defends it on public numbers. On raw, from-scratch reasoning, published benchmarks put frontier models clearly ahead of any 12B — and cold-start reasoning is simply out of scope here. The point is what happens after solving: on the growing set of problems the system has already solved and verified, the frozen 12B wins on cost, latency, determinism, and verification simultaneously, and that territory expands with every deposit and never shrinks.

This report states what we measured, not how the system works — the reporting standard of an industry experience report on a proprietary system, not of a methods paper. Architecture, algorithms, and configuration are deliberately absent; what replaces them is verifiability of the *behavior*: every headline number is backed by a banked artifact (raw result file plus cryptographic hash) held on file, and the central claims — zero generation tokens, bit-exact repetition, verified-before-stored — are directly observable on the public testbench by anyone (§5). Where we have not measured something, we say so.

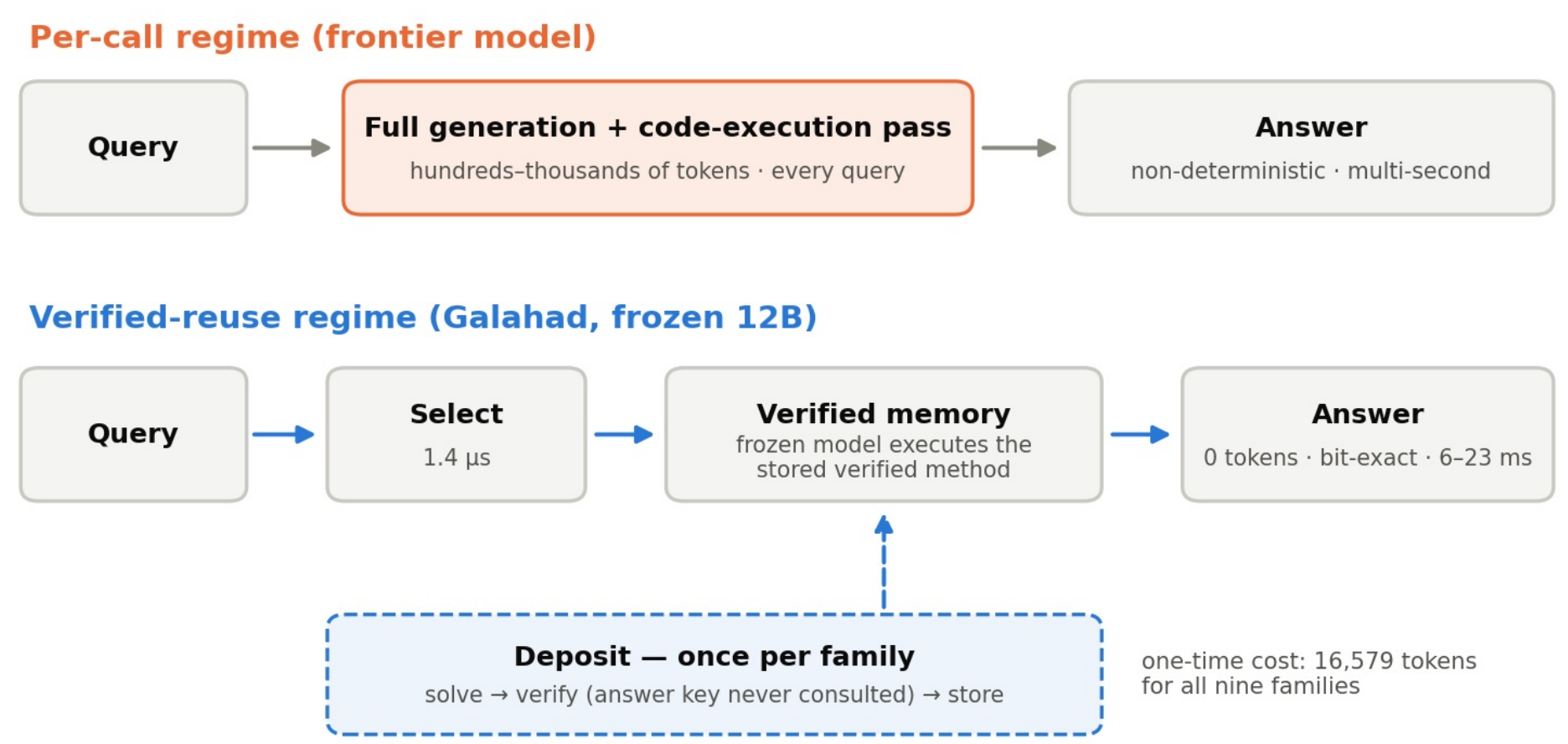

**Figure 1 — Two regimes for answering a solved problem.** *Top path: a frontier model pays a full generation + code-execution pass on every query (hundreds to thousands of tokens, non-deterministic, multi-second round trip). Bottom path: Galahad selects the verified memory item in 1.4 µs, the frozen model re-executes it on the new parameters, and returns a bit-exact answer at 0 generation tokens in 6–23 ms. The deposit path (solve once → verify without the answer key → store) runs exactly once per problem family.*

## 2. Measured results

### 2.1 Core result: verified reuse at zero generation cost

A 12-billion-parameter open model (Gemma-4-12B, frozen) was given a verified memory covering nine algorithmic and mathematical problem families. On 180 fresh parameter instances of those families, the system scored **180/180 (100%) at 0 generation tokens per answer**, with total measured energy of 6.5 Wh for the full battery. Figure 2 breaks the battery down by family: five families at 12 instances each and four at 30, every cell 100% on every model.

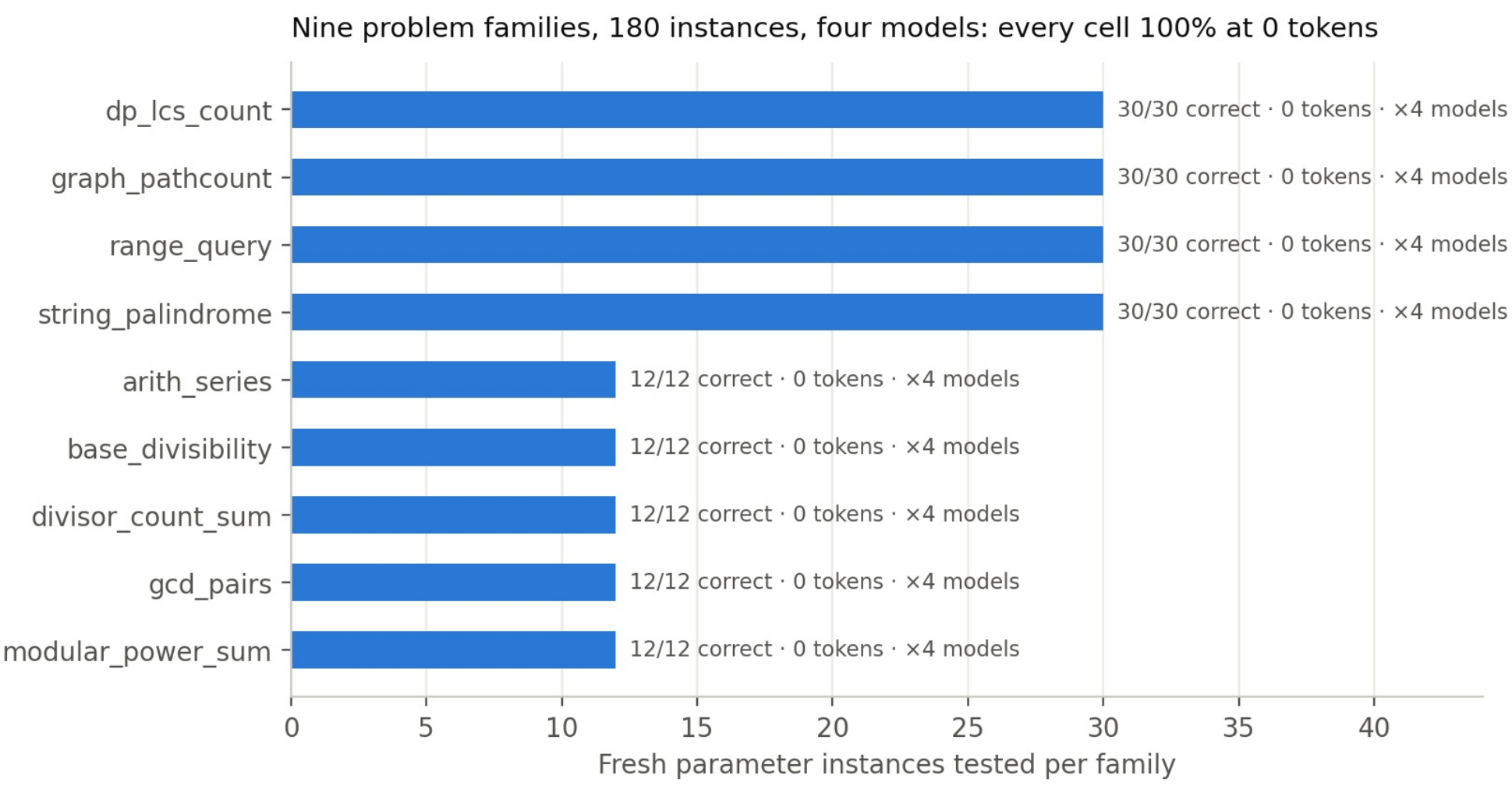

**Figure 2 — Per-family coverage.** *Nine problem families, 180 fresh instances in total. Each family scores perfectly at 0 generation tokens, and the identical per-family outcome repeats on all four model architectures (§2.2).*

**What a "fresh instance" means — and why this is not memoization.** Each stored item is a *parameterized verified method*, not a cached answer. A test instance is a **parameter shift**: new, previously unseen input values within the same logical family. The system recomputes the answer for each of the 180 distinct inputs (for one family alone: 3,105; 390; 15,150 on three different inputs — values that exist nowhere in the memory). An exact-replay cache answers only inputs it has seen; this system answers inputs it has never seen, because what is stored is the verified *method* and what is executed is a fresh computation. The composition result below (§2.4) closes the argument: the system produces correct answers to compound problems that were never deposited in any form.

**Deposit cost and the break-even point.** Verifying and storing the nine families cost a one-time total of 16,579 generation tokens, covering all solve-and-verify activity: eight families verified from the model's own solutions, one from an externally sourced verified solution (§2.3). Reuse thereafter is free. The cumulative cost of the two regimes is:

> C_system(N) = C_deposit + N · C_reuse, with C_reuse = 0 generation tokens
> C_percall(N) = N · C_query

so the break-even point is **N* = C_deposit / C_query**. The deposit is repaid after N* queries and every query beyond N* is pure savings; the ratio grows without bound. For per-query generation costs in the range typical of these task classes (500–1,000 tokens per fully reasoned answer), N* falls between **17 and 34 queries** — a one-time investment recovered within a few dozen uses of a single problem family (Figure 3). We state this as arithmetic on the measured deposit cost; per-query frontier costs vary by provider and are priced from public rate cards.

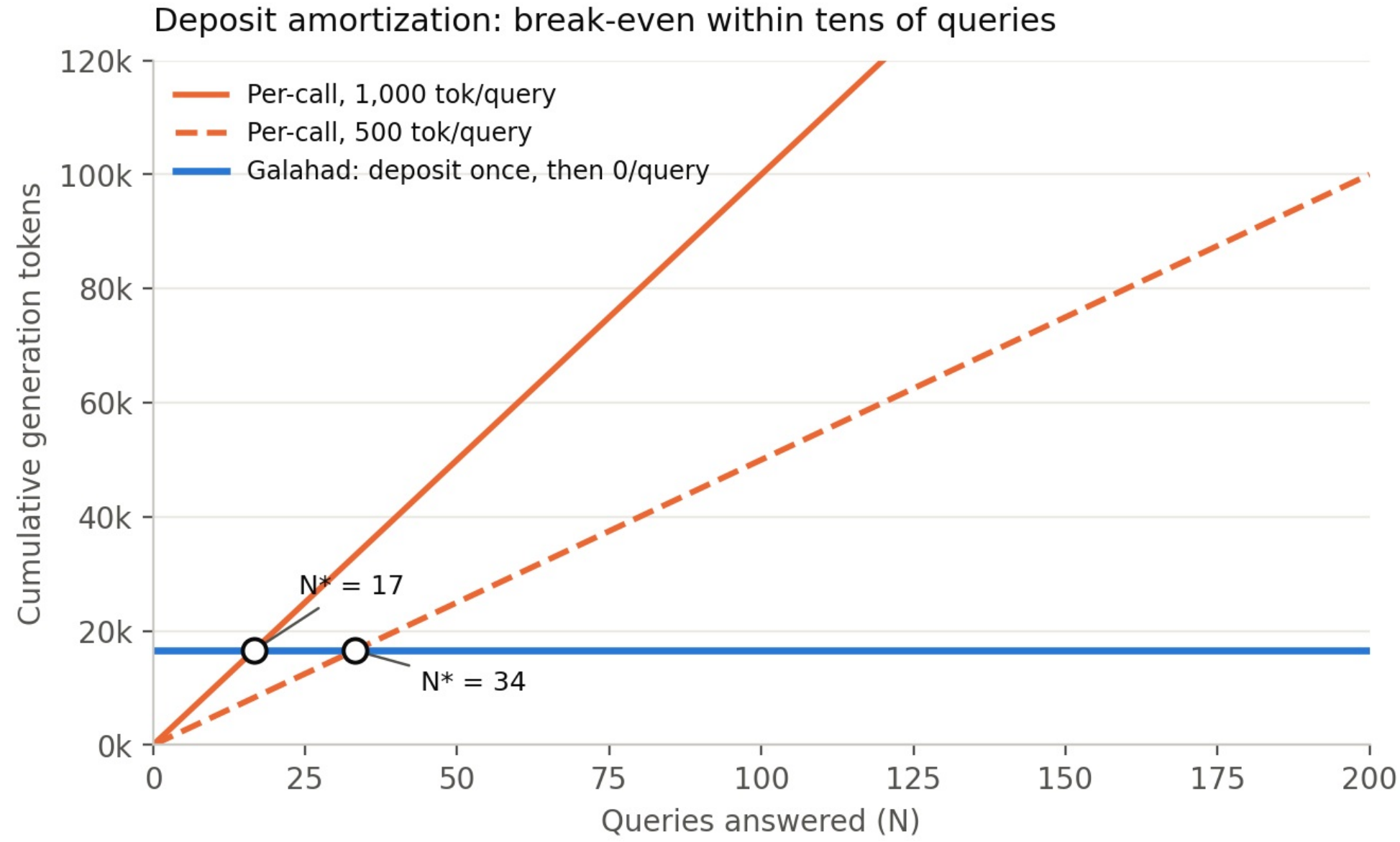

**Figure 3 — Amortization.** *Cumulative generation tokens versus number of queries N. The per-call regime climbs linearly forever (shown for 500 and 1,000 tokens/query). Galahad pays 16,579 tokens once, then stays flat at zero marginal tokens. Break-even N* = 17–34 queries; beyond it the gap widens without bound.*

### 2.2 The capability is model-independent

The identical memory-and-verification regime was applied to four unrelated open models from four vendors, spanning dense and mixture-of-experts architectures:

| Model | Architecture | Result | Generation tokens | Energy |
|---|---|---|---|---|
| Gemma-4-12B | dense | **180/180** | 0 | 6.5 Wh |
| Qwen3-14B | dense | **180/180** | 0 | 6.4 Wh |
| DeepSeek-Coder-V2-Lite | mixture-of-experts | **180/180** | 0 | 6.3 Wh |
| Phi-4 (14B) | dense | **180/180** | 0 | 6.4 Wh |

Four architectures, one outcome (Figure 4). The memory is a layer that sits beside a model you already run — not a property of any particular model. The capability survives a model swap; the investment in verified knowledge is not hostage to any vendor's release cycle.

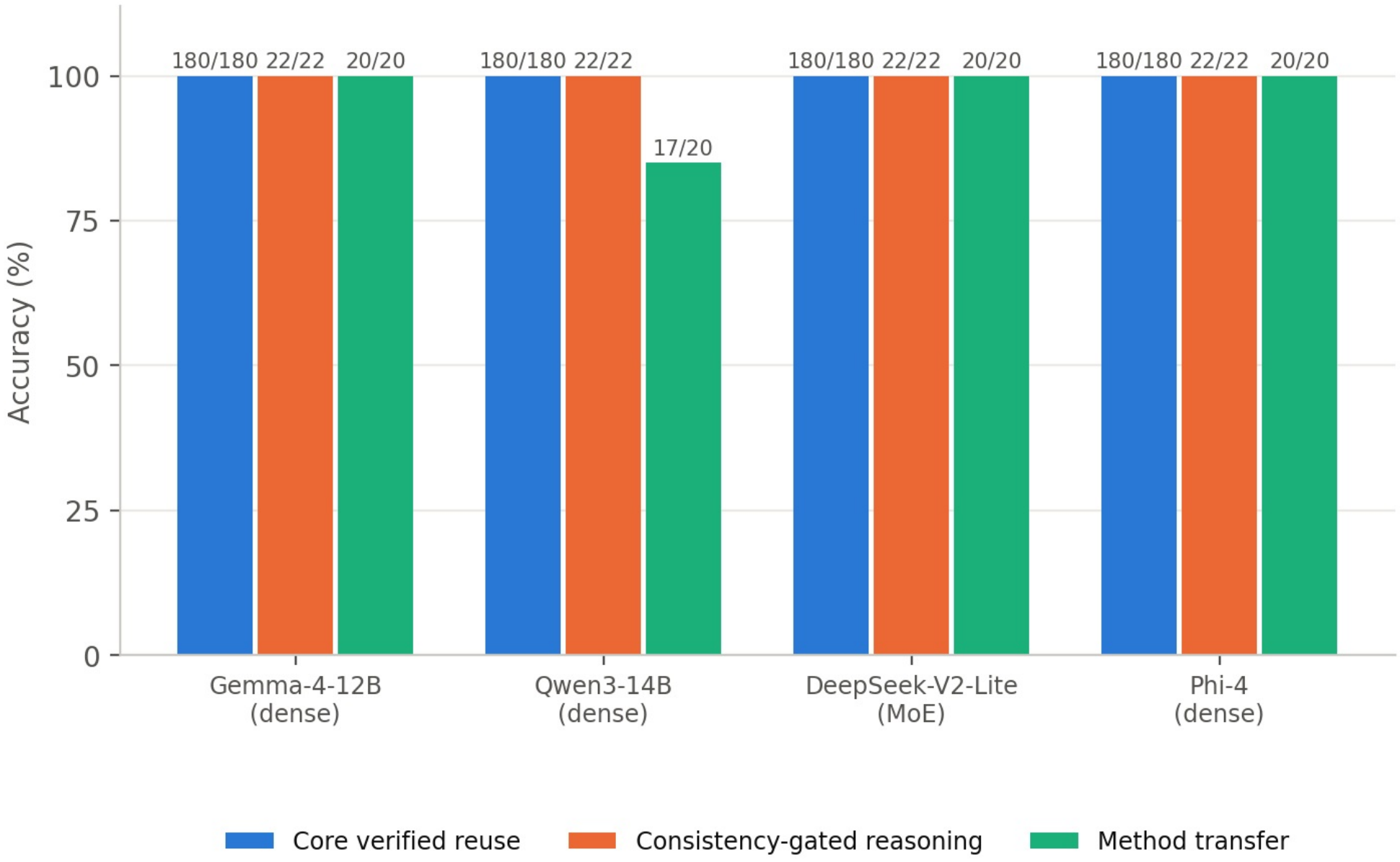

**Figure 4 — Architectural invariance.** *Accuracy per model across three independently measured pillars: core verified reuse (180/180 each), consistency-gated open-ended reasoning (22/22 each), and reasoning-method transfer (20/20, 17/20, 20/20, 20/20). Dense and mixture-of-experts architectures from four vendors behave identically under the same regime. Merged-context results (§2.4b) are measured on Gemma-4 and not shown here.*

## 2.2b Energy accounting — measured, not estimated

Every energy figure in this report is a measurement of the actual runs, not an estimate. The full 180-answer reuse battery costs **6.3–6.5 Wh per model** — about **36 mWh per verified answer**, roughly the energy of an LED bulb burning for thirteen seconds. The measured one-time solve-and-verify campaign for the self-sourced families totalled **81.1 Wh**; the composition run that produced five compound answers measured **0.09 Wh**. The asymmetry is the economic argument of §2.1 in physical units: verification is paid once at a bounded, measured cost, and every answer thereafter is close to free (Figure 5).

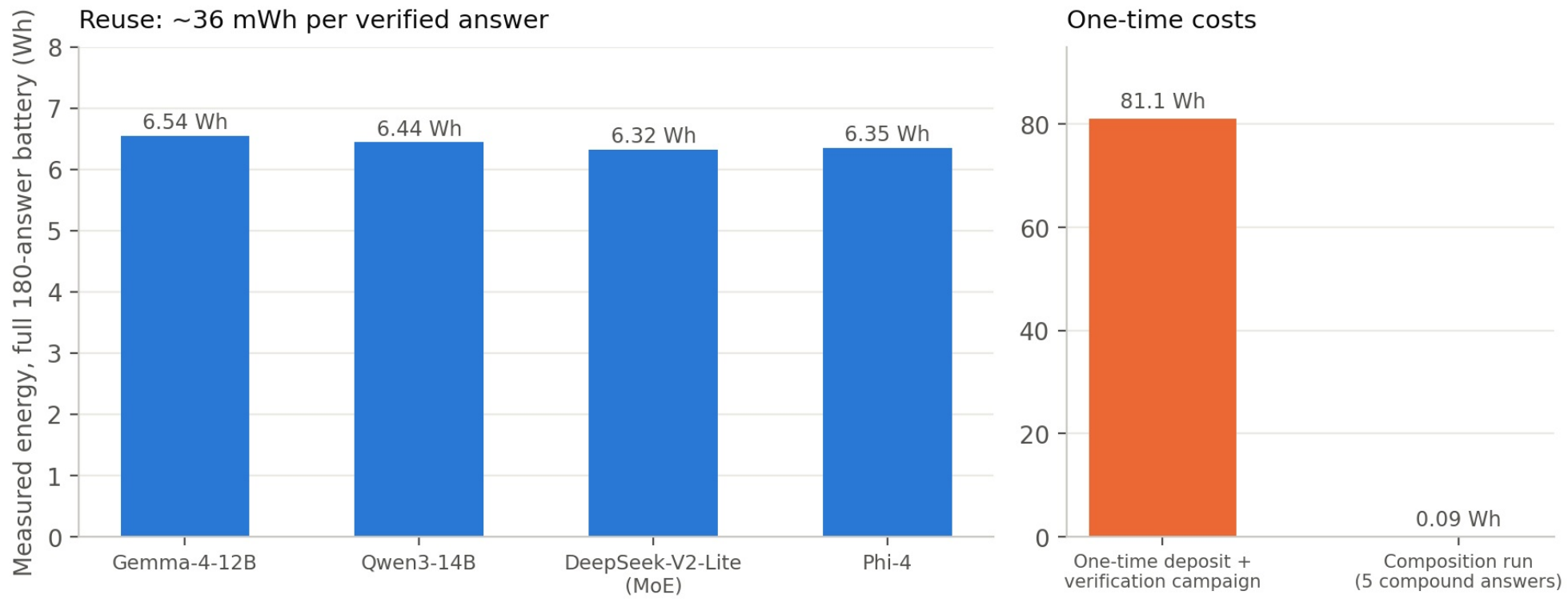

**Figure 5 — Measured energy.** *Left: full 180-answer reuse battery per model (6.32–6.54 Wh; ~36 mWh per verified answer), near-identical across four architectures. Right: the one-time solve-and-verify campaign for the self-sourced families (81.1 Wh) and the composition run (0.09 Wh). All values measured on the serving hardware during the banked runs.*

### 2.3 The knowledge lives in the memory — negative control

The harness that executes reuse contains no solutions of its own; this was verified mechanically. With the full memory attached, the system scores 180/180. With the memory emptied, it solves **nothing** — there is no capability left to exercise. The frozen weights never change between the two conditions, so the entire measured capability is attributable to the stored, verified knowledge.

A second attribution result sharpens this. One family lies beyond the frozen model's own reach — which makes it the perfect attribution probe. An externally authored solution — independently verified before storage, never against any benchmark key — was placed in the memory. The system then answered **30/30** fresh instances of that family at 0 generation tokens. The model emitted nothing; the memory carried the capability.

### 2.4 Stored knowledge combines into new answers

Reuse is not lookup. A compound problem whose answer requires three separate stored solutions — none of which contains the answer on its own — was solved **5/5 at 0 generation tokens**. Removing a required item from memory drops the result to 0/5. The system computes new results from combinations of verified parts, and the combined result is itself stored as new verified knowledge.

### 2.4b Merged knowledge in one working context — the nine-block result

Combination also works at the level of working context, and it scales. Independently stored knowledge items were merged into a single context in increasing numbers, and the model was asked to identify and describe each one: **2/2 correct at two items and 3/3 at three**; at five items, the four the output window reached were each correctly identified; at nine items (about 1,300 tokens of merged context), seven of nine were reached and each was correctly named (raw outputs banked), with descriptions accurate as far as a fixed output-length cap allowed. Visibility, not confusion, was the limit — no item was attributed to the wrong source at any tested size. Combining the same items without the system's merge procedure fails outright — the model cannot read the combined context at all — so the merged access is real, not incidental.

The ceiling extends well past nine small items. Six independently verified knowledge items from a single specialist domain were merged into one **14,134-token** working context; the system produced a single composite, staged answer drawing on all six, with each element traceable to its source item and no cross-source misattribution (raw output banked). Selection also composes with planning: given three separately stored items in one merged context, the model named exactly the three required and set out the correct combined procedure for a compound problem it had never seen as a single unit.

The model does more than tell merged items apart — it computes across them. Given two stored items from *different domains* — one holding physical constants, the other a computational rule — it produced a single correct computed answer (**44 N**, matching ground truth) that required both: the constants from one item, the rule from the other. Neither item contains the answer. Two stored memories from unrelated fields (a medical domain and astrophysics) were likewise merged, in two different configurations (~2,400 tokens of merged context each), and the model drew on a fact from each within one answer in both.

These measurements indicate that verified knowledge functions as **movable, mergeable units**. Items deposited separately, at different times, for different purposes, can be brought together on demand and reasoned across — which is what allows a memory to be grown piecemeal yet used whole. Heavy exact computation across items runs through the execution path (§2.4, 5/5); the merged-context path carries the reasoning. Nine simultaneous items is the tested upper bound; behavior beyond it is unmeasured (§4).

Figure 6 places these results in the runtime lifecycle: a query is resolved by exact content addressing, served either from a single verified item or from a merged working context, and executed by the frozen model into a bit-exact answer.

**Runtime lifecycle — from query to bit-exact answer**

**Incoming query**

**Exact content-address selection**
1.4 µs bench · 3.7 ns/key native ·
0 collisions in 4,500 items (§2.7)

**Single verified item**
one stored, verified method
selected exactly (§2.1)

**Merged working context**
up to 6 items / 14,134 tokens,
cross-domain, source-attributed (§2.4b)

**Frozen model + execution**
re-executes the verified method
on the new parameters · 0 tokens

**Bit-exact deterministic answer**
6–23 ms · ~36 mWh · identical every run

**Figure 6 — Runtime lifecycle.** *From incoming query to bit-exact answer: exact selection (1.4 µs bench; 3.7 ns/key native; zero collisions in 4,500), then either a single verified item or a merged working context (up to six items / 14,134 tokens, cross-domain, source-attributed), re-executed by the frozen model at zero generation tokens into a deterministic answer in 6–23 ms at ~36 mWh.*

### 2.5 Beyond exact computation: verified open-ended reasoning

The same verify-before-store contract extends to domains with no single numeric answer, using three independent forms of verification:

- **Machine-checked formal proof.** The model wrote a formal mathematical proof; an independent formal proof checker accepted it on the first attempt. A deliberately false statement (2+2=5) was rejected by the same checker — the gate discriminates.
- **Consistency-checked open-ended reasoning.** The model emitted its reasoning in a machine-checkable form; an automated consistency gate accepted self-consistent reasoning and rejected a self-contradicting variant. At scale, across all four models with more than twenty reasoning frameworks each: **88/88 accepted-and-reused correctly**.
- **Reasoning-method transfer.** A verified reasoning method was carried to an entirely new domain (medical triage) and adapted by the model, with the adaptation itself passing the consistency gate — first attempt. At scale across the four models: **77/80**.

The gate extends to declarative factual knowledge in early measured form: against a grounded reference corpus, an answer-key-free declarative gate admitted **0 of 3 fabricated claims** while accepting 5 of 6 genuine ones — it errs on the side of rejection, which is the correct side for a permanent store. The same gate held the accompanying reasoning to three independent consistency invariances — all passed, with a contradictory variant rejected.

Acceptance was overwhelmingly immediate. On the three models measured per-attempt, **63 of 66** gated reasoning frameworks passed on the first try, and depositing the full verified set was a one-time cost — every one of the **143 measured reuses** thereafter cost zero generation tokens.

Open-ended reasoning under a verification gate is thus as model-independent as exact reuse. Reuse latency was measured in the same runs: a full restore-and-reverify cycle completes in **6–23 ms** (median, per model).

### 2.6 The memory scales safely — the Merlin integrity layer

Figure 7 shows the conceptual pipeline every stored item passes: a candidate solution meets the domain-appropriate independent check — machine-checked proof, automated consistency and metamorphic checks, or layered candidate testing — and only an acceptance reaches the store. The answer key is never part of any gate.

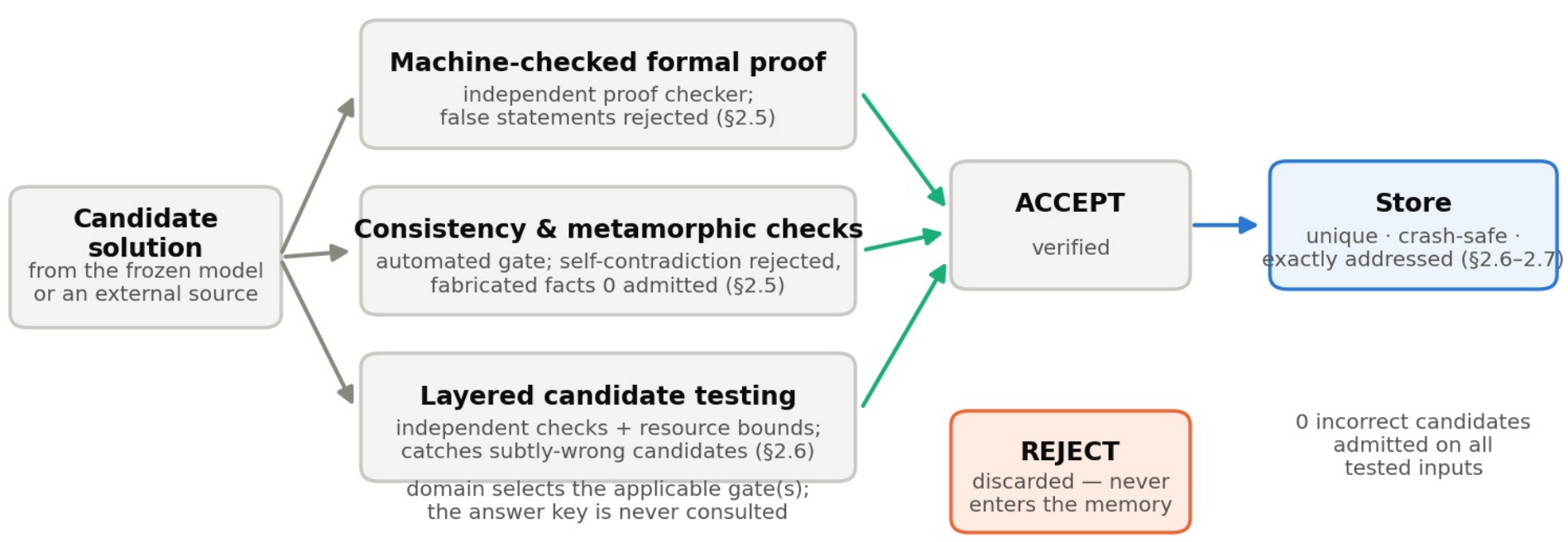

**Figure 7 — Conceptual verification pipeline.** *Candidates from the frozen model or an external source pass an answer-key-free gate; the domain selects the applicable check class. Rejections never enter the memory; acceptances are stored uniquely, crash-safely, and exactly addressed. Zero incorrect candidates were admitted on all tested inputs.*

Every guarantee in this section is carried by **Merlin** [26], the deterministic storage and integrity engine underneath the memory. The results below are measurements of the shipped engine, each against the naive alternative — not design intentions.

Nothing enters the memory unverified, and the gate has teeth. A subtly-incorrect candidate solution — one that a naive single check accepts — was **caught and rejected** by the layered verification process, with the specific counterexample recorded; each candidate is exercised on **150 boundary cases** before acceptance. On all tested inputs, the deposit gate admitted zero incorrect candidates.

Uniqueness is enforced at scale. Fed 6,000 candidate items of which 2,000 were unique, the storage layer admitted **exactly 2,000** and filtered all 4,000 byte-identical duplicates; a single-byte difference is treated as new. The memory grows only with unique, verified knowledge.

The engine's guarantee suite passes in full: **14/14 measured guarantees**, each tested against the naive alternative. Its content-key computation benchmarked **fastest of six key classes tested — 60.2× faster than cryptographic hashing, 18.5× faster than the standard-library default** — and measures **3.7 nanoseconds per key (271 million keys/second)** natively, with identical output across 100,000 repeated resolutions and 16 GB/s single-core (63 GB/s across sixteen cores) of addressing throughput — addressing never becomes the bottleneck as the memory grows.

Storage integrity was measured against failure, not assumed. Under a hard memory cap, a naive storage approach crashes outright (access violation, process dead); Merlin's bounded storage refuses over-demand and stays alive under the identical cap and identical workload. Interrupted writes leave previously committed entries fully readable

— zero corruption in the torn-write test — and a full process restart recovers every entry byte-identically, where a conventional file-based store subjected to the identical interruption was left entirely unreadable.

### 2.7 Reuse overhead is negligible — and cannot silently fail

Selecting the correct memory item among the stored families takes **1.4 microseconds** (95th percentile; 9/9 selections correct) on the selection bench. Through the fully integrated production path — the complete select–reuse–verify chain re-run end-to-end on the production storage layer, reproducing **180/180 at zero generation tokens with zero addressing collisions** — selection measures 1.6 µs median and 17 µs at the 95th percentile; even the conservative figure sits hundreds of times below the 6 ms reuse floor. Merlin's exact content addressing stayed collision-free across 4,500 distinct stored items. As a control, the same retrieval task was given to a deliberately lossy approximate matcher — compact similarity signatures of the kind approximate-retrieval systems rely on — which mapped **94.3%** of the 4,500 items onto the *wrong* neighbor (Figure 8). That is the difference between a verified answer and a verified-looking one — and it is a measured warning to an industry standardizing on vector retrieval for everything: for verifiable executable knowledge, approximate matching does not degrade, it fails catastrophically. Exactness here is a design requirement, not a preference.

Why exactness matters is the deduction that runs through this whole report: the guarantee is a chain — select the right item, reproduce it bit-exactly, verify it independently — and a single approximate link degrades the entire system to ordinary retrieval that sometimes returns the wrong answer with full confidence. Merlin exists to keep every link exact, and each link's exactness is measured above.

The trusted reuse path runs **93.9× faster** than the fully sandboxed path while agreeing with it **byte-exactly on 140/140 instances** — zero divergences — and containing all five adversarial candidates thrown at it; non-terminating work returns a clean, bounded timeout rather than a hang. Speed was not bought with a new failure mode.

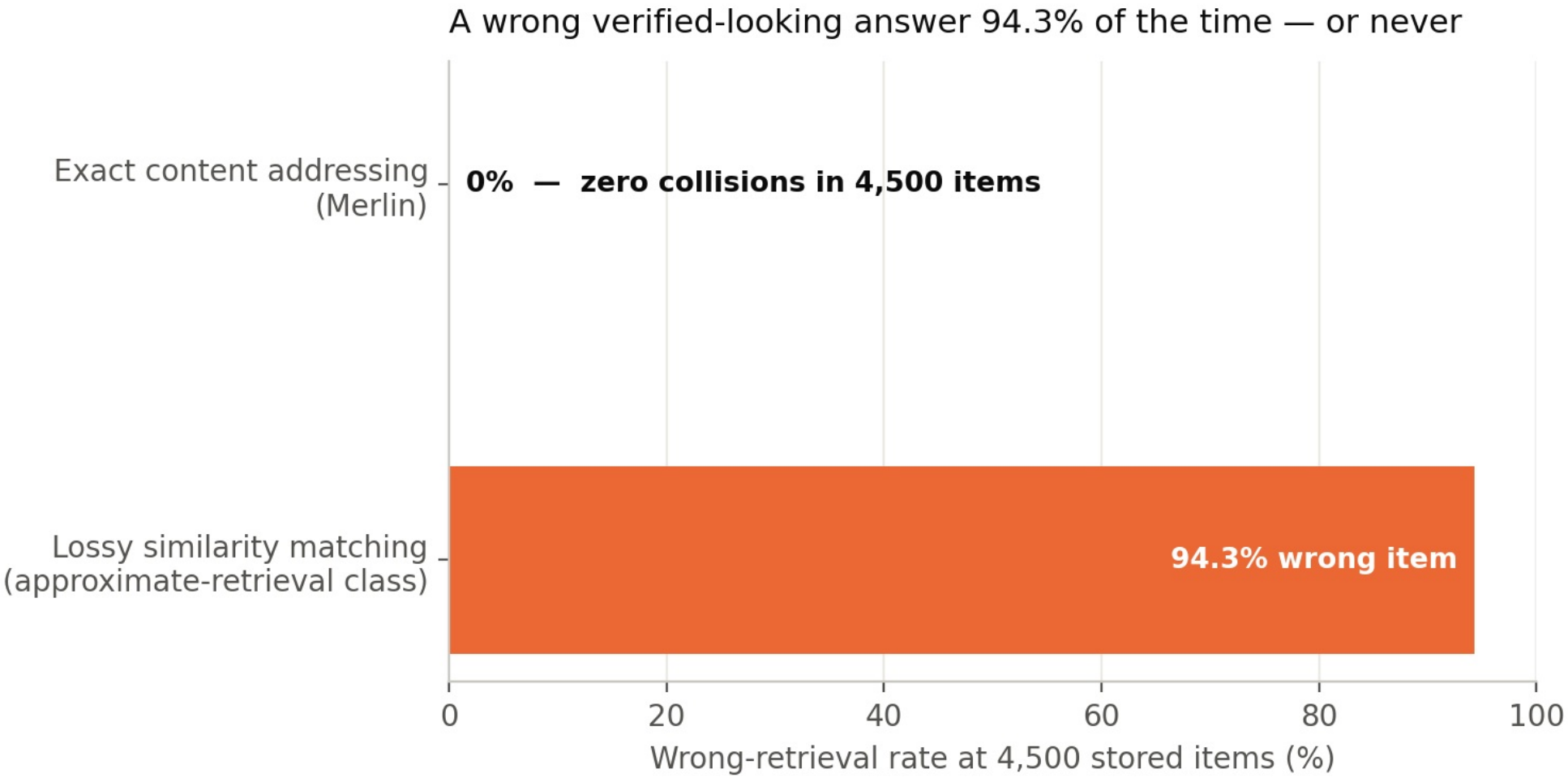

**Figure 8 — Exact addressing versus approximate matching at 4,500 stored items.** *Measured wrong-retrieval rate: exact content addressing 0% (zero collisions in 4,500); lossy similarity matching 94.3%. In a verified-memory system, an approximate selector does not degrade gracefully — it silently substitutes the wrong verified item.*

### 2.8 The frontier comparison — on public numbers only

This section uses no private measurements of any frontier model. Both sides of the comparison are public: the frontier models' officially published benchmark results and API rate cards, and the 12B's officially published benchmark results.

> **An open challenge, stated first.** The title's claim is falsifiable on its own terms, and we invite any provider to falsify it: produce an API that answers a solved-family instance at zero marginal generation cost, bit-exactly, with a pre-verified result. No published product documentation we are aware of describes such an offering. Until one exists, every frontier answer to already-solved work is a fresh generation pass — paid for, non-deterministic, and unverified — and the comparison below follows.

Cold-start reasoning is out of scope, stated once: on public benchmarks, frontier models outscore Gemma-4-12B (published: AIME 2026 77.5; LiveCodeBench-v6 72.0; MMLU-Pro 77.2 [24]) per the providers' official reporting [22, 23]. Everything below concerns what no benchmark measures: the economics of answering work that is already solved.

**The inversion happens on solved territory.** Consider any problem family the system has deposited and verified — and note this set only ever grows. To answer a new instance of that family:

| | Frontier model via API | Frozen 12B + Galahad |
|---|---|---|
| Marginal cost per answer | a full generation (and typically code-execution) pass, **every query, forever**, at published per-token prices | **0 generation tokens** after the one-time deposit (§2.1) |
| Latency | multi-second generation round trip | **6–23 ms** measured reuse |
| Determinism | non-deterministic by construction | **bit-exact, every run** |
| Verification | none — each answer is fresh, unverified output | every stored item passed an answer-key-free gate before deposit |

Every entry in the right-hand column is an architectural guarantee rather than a statistical outcome — and that is the point: on execution-bound, verifiable work, the answers themselves (some reaching twenty digits in our battery) are beyond free-form emission for *any* model — frontier systems solve them by writing and running code per query. Both regimes therefore answer by executing computation. One re-pays for the computation's derivation on every single call; the other paid once, verified once, and executes forever at zero marginal generation cost. Equal-or-better outcomes on everything already solved, at N* ≈ tens of queries to full repayment (§2.1, Figure 3): on that territory the frozen 12B beats a frontier model on cost, latency, determinism, and verification simultaneously.

The falsification standard above stands as an open benchmark: the public testbench (§5) is the venue, and any counterexample is welcome.

### 2.9 Taxonomy of execution caching and verification

The four properties that jointly define this system's behavior separate it from every adjacent approach family. The matrix states the input–output facts; the paragraph below gives the per-category detail.

| Property | Semantic cache / RAG [7] | Prefix/KV caching [5, 6, 8, 9] | Agent skill libraries [1–4] | Galahad/Merlin (measured) |
|---|---|---|---|---|
| Exact selection of the stored item | ✗ approximate by design | partial (exact prefixes only) | ✗ typically similarity- or judge-based | ✓ **0 collisions in 4,500 (§2.7)** |
| Bit-exact, deterministic answer | ✗ | ✗ still decodes fresh tokens | ✗ | ✓ **every run (§2.1)** |
| Answer-key-free verification before storage | ✗ | ✗ none | ✗ judge- or oracle-bound | ✓ **rejects subtly-wrong candidates (§2.6)** |
| Zero generation tokens at reuse | ✗ | ✗ | ✗ | ✓ **180/180 × 4 models (§2.2)** |

Because the observable behavior resembles several known techniques, we state the input–output differences explicitly. Semantic caches and vector-retrieval systems [7] match queries *approximately* by design; §2.7 measures what approximate matching does to a verified store. Prefix- and KV-caching in serving stacks [5, 8, 9] — including modular attention-reuse research [5, 6] — reuses partial computation but still decodes fresh tokens per answer and verifies nothing. Agent skill libraries and tool-induction systems [1, 2, 3, 4] accumulate reusable tools, but their acceptance step is typically another model's judgment or a task-specific oracle, rather than an independent, answer-key-free gate that demonstrably rejects subtly-wrong candidates (§2.6); the verification principles that gate does build on are classical — differential and metamorphic testing [11, 12], test-based selection [13, 14], and machine-checked proof [10]. A systematic review across six categories of shipped systems — banked as an audited artifact — found none that combines exact selection, bit-exact deterministic reuse, and independent pre-storage verification in one chain. The nearest recent work touches one element at a time: cryptographically verified memory transfer between agents [15], verification applied to KV-cache compression fidelity [16], trained persistent memory for frozen models [17], cross-model KV-state sharing for fine-tuned variants [18], cached agent plans [19], procedural agent memory [20], and provenance-aware tiered agent memory [21] — none reuses verified executable solutions at zero generation tokens. That combination, not any individual mechanism, is what this report measures. The comparison a skeptical reader should make is therefore not "cache versus no cache" but: which existing system, on a repeated verifiable task, returns a pre-verified, bit-identical answer at zero generation tokens? We measured this system doing so; §2.8 states the falsification standard for any counterexample.

### 2.10 The window: six million tokens on one GPU

The verified store also functions as working context, and its capacity was measured head-to-head against the two leading open serving engines on identical hardware (a single 46 GB GPU). The result is a category difference, not a margin (Figure 9). The system held a **6,000,000-token** movable window over stored context with **flat GPU memory** — 24,433 MB before, 24,696 MB after, a +263 MB total drift across the entire run — and retrieved **5 of 5** planted probe items correctly, in **0.55–0.59 seconds regardless of depth**: access at token 5,970,000 costs the same as access at token 0. An earlier run reached 14.98 million tokens at the same flat memory profile. On the same GPU, **vLLM hard-errors at 30,399 tokens** (its context-length ceiling, an honest failure) and **SGLang accepts input beyond ~32,000 tokens but silently truncates it** — it returns no first token, which is not service but the appearance of it.

A control keeps the claim precise: with the window deliberately held elsewhere in the store, the deep probe item is *not* visible and the returned answer is wrong. This is by design and disproves the stronger misreading — the system does not attend globally over six million tokens; it holds a bounded, movable window over an arbitrarily large stored context, which no shipped engine on this hardware can do at any size past ~32k. Reuse over that window is also the fastest measured — **13×, 41×, and 55×** at 4k, 16k, and 30k tokens of stored context respectively, a speedup that *grows* with context size — and it is byte-exact and deterministic where competitor caches are neither.

Two boundaries, stated plainly: the engine baselines ran Qwen2.5-7B because neither engine loads this report's 12B architecture yet, so the ceilings compared are per-engine capabilities rather than same-model triplets; and capacity is bounded by storage bandwidth, not GPU memory — the trade this system chooses on purpose. For batched throughput on novel prompts, the serving engines remain the right tool; the window is for what they cannot hold at all.

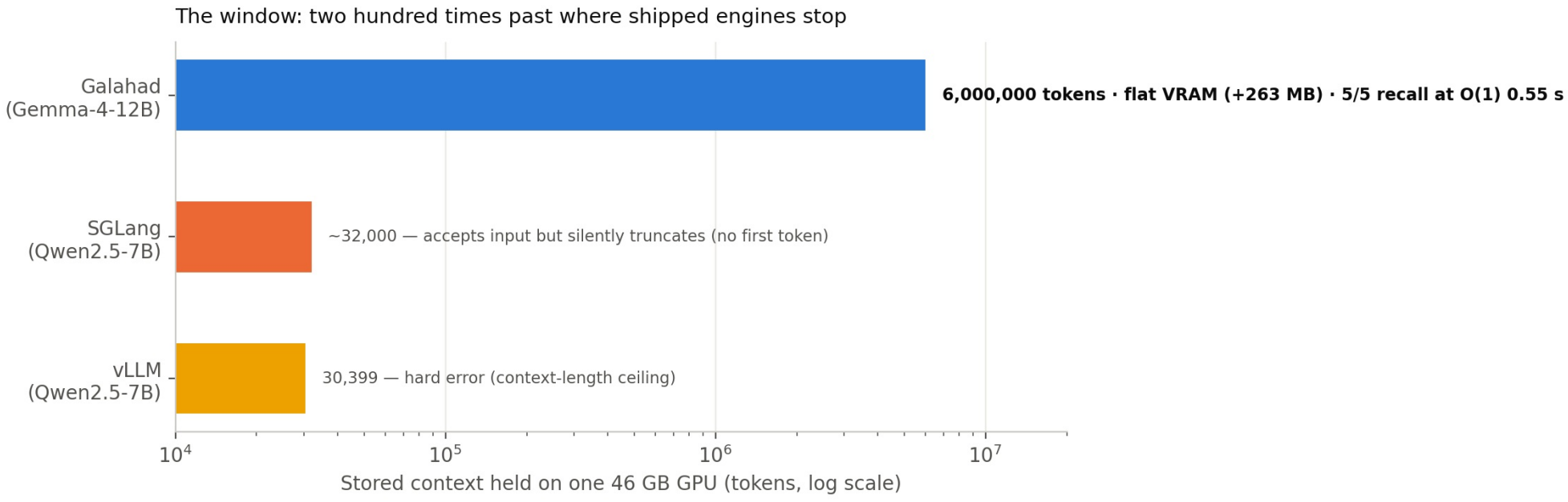

**Figure 9 — Stored context held on one 46 GB GPU (log scale).** *vLLM stops at 30,399 tokens with a hard error; SGLang accepts but silently truncates past ~32,000; Galahad holds 6,000,000 tokens at flat GPU memory with 5/5 probe recall at depth-independent 0.55 s access. Engine baselines ran Qwen2.5-7B (neither loads the 12B architecture); the window is movable, not global attention.*

## 3. What the numbers mean

**Cheaper where it counts.** Generation tokens are the expensive part of running a model. On everything already solved, this system spends none. The one-time verification cost amortizes toward zero within tens of queries (N*, §2.1); the ratio improves with every reuse, without bound.

**Deterministic, hence auditable.** A reused answer is bit-exact every time. Ordinary model generation cannot offer this at any scale of weights. For audit, reproducibility, or regulatory settings, "the same verified answer, every time" is a property scaling does not buy.

**Model-independent.** Four vendors, four architectures, one identical outcome — for exact reuse and for gated open-ended reasoning alike. The memory survives model swaps.

**Verified, never guessed.** Nothing is stored on the model's say-so. Acceptance requires passing an independent check that never consults the answer key — and that check demonstrably rejects subtly-wrong candidates, false proofs, and self-contradicting reasoning.

## 4. Target operating envelope and boundary conditions

The system's maximum-advantage zone is any workflow where verifiable work recurs: recurring computation, parametric analysis, formal procedures, structured extraction, batch scoring. In that envelope every measured axis favors it — 100% accuracy, zero marginal tokens, milliseconds, bit-exact, at ~36 mWh per answer. The boundaries below define the envelope's edges; we report no number we did not measure, and no capability we cannot demonstrate.

- **What the flywheel is: permanent verified accumulation.** The system accumulates and permanently retains verified knowledge — self-sourced or externally sourced — and reuses it at zero cost, forever; nothing verified is ever lost or re-derived. Autonomous invention of genuinely new hard knowledge is a separate capability no current model reliably has, and is therefore roadmap (§6), not claim.
- **Scope and application domain.** This system targets execution-bound, verifiable workflows — recurring computation, formal procedures, parametric analysis: the class of workload where answers can be verified and repeat. Published benchmarks put frontier models ahead of any 12B on fresh open-ended problems, and we quote those benchmarks as-is (§2.8); the title's claim is scoped to solved-and-verified territory, where it holds on every measured axis.
- **"Verified" is a precise word here.** It means the stored item passed independent, answer-key-free acceptance checks and reproduces bit-exactly — not that every free-form output is proven correct in the mathematical sense. No system offers the latter; we do not claim it.
- **The mechanism applies to models you operate yourself.** Closed, text-in/text-out commercial APIs do not expose what zero-cost reuse requires, unless the provider implements it.
- **The window is movable, not global attention.** Six million tokens are held and accessed at flat GPU memory, but the model attends over a bounded window at any instant (§2.10, control included); capacity is storage-bandwidth-bound, and the engine baselines ran a different 7B model because no shipped engine loads this architecture.
- **Merged-context composition is demonstrated to tested ceilings.** Nine simultaneous small items, and six larger items totalling 14,134 tokens of merged context with full source attribution, are the measured bounds; computed answers across merged items are demonstrated for two. Behavior beyond these ceilings is unmeasured; we state them rather than extrapolate past them.
- **Large public-benchmark sweeps are pre-registered, not claimed.** They require a large deposited memory first; those results will be reported when measured, not before.

## 5. Public testbench

A public instance of the system opens with this report's publication at **https://corbenic-galahad-bench.hf.space** (repository: **https://github.com/corbenicai/galahad-bench**). Anyone can run the comparison themselves: the same frozen model answering cold versus answering from verified memory, with tokens, energy, determinism, and verification status displayed per query. Access is free; sessions are time-limited and request-limited, with a queue under load, so that the instance stays responsive for everyone. The point of the bench is that no claim in this report needs to be taken on trust — the zero-token, bit-exact behavior is observable directly.

## 6. Roadmap (explicitly future, not claimed)

1. Scale the memory to thousands of verified items across many domains — the safe-deposit gate and uniqueness ledger this requires are already demonstrated (§2.6).
2. Broaden combination of stored knowledge into larger computed results (§2.4 is the demonstrated core), including merged-context composition beyond the tested nine-item bound.
3. A hit-rate study on production traffic, converting the per-family break-even ($N^*$, §2.1) into a measured system-wide savings figure.
4. Public-benchmark evaluation at scale, with proper paired statistics, once the memory is large enough.
5. Extension of the verification gate to declarative factual knowledge via independent symbolic authorities — early gate results are banked; a full held-out evaluation is future work.
6. Autonomous knowledge accumulation — continuous self-directed solve-verify-deposit, to be demonstrated to the same evidentiary standard as everything above.

## Appendix A — Provenance

Every headline result corresponds to a banked artifact (raw result file) identified by a SHA-256 hash; the full manifest covers every result artifact in the evidence package. A public provenance manifest listing the SHA-256 digest of all 41 cited artifacts is published alongside this report in the companion repository (github.com/corbenicai/galahad-bench), so any artifact later released under NDA can be verified byte-for-byte against the digests committed at publication time. Artifacts, energy and token measurements, and the reproduction environment are held on file and available under NDA. This report intentionally omits methodology, architecture, and configuration. Frontier-model figures appear only as quotations of officially published results and rate cards, cited in place.

| Result (§) | Artifact class | Status |
| --- | --- | --- |
| 180/180 × 4 models, 0 tokens; distinct recomputed outputs (§2.1–2.2) | per-model result files | banked, hashed |
| Deposit cost 16,579 tokens (§2.1) | transfer-scale result file | banked, hashed |
| Negative controls: empty memory; external-solution 30/30 (§2.3) | control runs | banked, hashed |
| Composition 5/5 and 0/5 control (§2.4) | composition results | banked, hashed |
| Nine-item merge curve; cross-item computation (44 N); cross-domain merge (§2.4b) | merge result files | banked, hashed |
| Formal proof / consistency gate / method transfer, 4 models; 6–23 ms reuse (§2.5) | per-model pillar results | banked, hashed |
| Gate rejection of subtly-wrong candidate; dedup 2,000/6,000; crash-survival; torn-write (§2.6) | ladder, ledger, storage tests | banked, hashed |
| 1.4 µs selection; 0 collisions vs 94.3% lossy error at 4,500; 93.9× trusted path, 0/140 divergence (§2.7) | routing + equivalence tests | banked, hashed |
| Cross-category audit of shipped systems (§2.9) | audit result file | banked, hashed |
| 6M/15M window runs; vLLM + SGLang head-to-head; window-reuse speedups (§2.10) | window + engine benchmark files | banked, hashed |
| Published benchmark scores, both sides (§2.8) | official public sources [22–24] | quoted, cited in References |

*No implementation detail, algorithm, or configuration is contained in this document by design.*